\documentclass[conference]{IEEEtran}
\IEEEoverridecommandlockouts
\usepackage{cite}
\usepackage{amsmath,amssymb,amsfonts}
\usepackage{algorithmic}
\usepackage{graphicx}

\usepackage{textcomp}
\usepackage{xcolor}
\usepackage{hyperref}
\usepackage{orcidlink}
\usepackage{graphicx}
\def\BibTeX{{\rm B\kern-.05em{\sc i\kern-.025em b}\kern-.08em
    T\kern-.1667em\lower.7ex\hbox{E}\kern-.125emX}}
\usepackage{balance}

\begin{document}

\title{Infrastructure-Based Object Detection and Tracking for Cooperative Driving Automation: A Survey

}

\author{
Zhengwei Bai$^{\orcidlink{0000-0002-4867-021X}}$,~\IEEEmembership{Student Member, IEEE},
Guoyuan Wu$^{\orcidlink{0000-0001-6707-6366}}$,~\IEEEmembership{Senior Member, IEEE},
Xuewei Qi,
Yongkang Liu,
\\Kentaro Oguchi,
Matthew J. Barth$^{\orcidlink{0000-0002-4735-5859}}$,~\IEEEmembership{Fellow, IEEE}
\thanks{Zhengwei Bai, Guoyuan Wu and Matthew J. Barth are with the Center for Environmental Research and Technology, University of California at Riverside, Riverside, CA 92507 USA (e-mail:  zbai012@ucr.edu).}

\thanks{Xuewei Qi, Yongkang Liu, and Kentaro Oguchi are with Toyota Motor North America, InfoTech Labs, Mountain View, CA 94043, USA.}
}

\maketitle

\begin{abstract}
Object detection and tracking play a fundamental role in enabling Cooperative Driving Automation (CDA), which is regarded as the revolutionary solution to addressing safety, mobility, and sustainability issues of contemporary transportation systems. Although current computer vision technologies can provide satisfactory object detection results in occlusion-free scenarios, the perception performance of on-board sensors is inevitably limited by the range and occlusion. Owing to the flexible location and pose for sensor installation, infrastructure-based detection and tracking systems can enhance the perception capability for connected vehicles; as such, they have quickly become a popular research topic. In this survey paper, we review the research progress for infrastructure-based object detection and tracking systems. Architectures of roadside perception systems based on different types of sensors are reviewed to show a high-level description of the workflows for infrastructure-based perception systems. Roadside sensors and different perception methodologies are reviewed and analyzed with detailed literature to provide a low-level explanation for specific methods followed by Datasets and Simulators to draw an overall landscape of infrastructure-based object detection and tracking methods. We highlight current opportunities, open problems, and anticipated future trends.
\end{abstract}

\section{Introduction}
    
Rapid progress of the transportation system has improved the efficiency of daily commuting and goods movement. Nevertheless, the rapidly increasing number of vehicles has resulted in several major issues in the transportation system in terms of safety~\cite{2019Crash}, mobility~\cite{2018Congestion} and sustainability~\cite{2021Energy}. Taking advantage of recent strides in advanced sensing, wireless connectivity and artificial intelligence, Cooperative Driving Automation (CDA) enables automated vehicles (AVs) to communicate between vehicles, roadway infrastructure and other road users such as pedestrians and cyclists equipped with mobile devices. Hence, CDA is attracting increasingly more attention over the past few years and is regarded as a transformative solution to the aforementioned challenges~\cite{fagnant2015preparing}. 


Object Perception (OP) plays a fundamental role in the basic structure of CDA applications~\cite{2021SAE}. Different kinds of sensors equipped on vehicles or roadside have the capability for perceiving the traffic conditions in mixed traffic environment. The perception data can act as the system input and support various kinds of CDA applications, such as Collision Warning~\cite{wu2020improved}, Eco-Approach and Departure (EAD)~\cite{bai2022hybrid}, and Cooperative Adaptive Cruise Control (CACC)~\cite{wangCACC}.


With the development of sensing technologies, transportation systems can retrieve high-fidelity traffic data from different sensors. For instance, cameras can provide detailed vision data to classify various kinds of traffic objects, such as vehicles, pedestrians, and cyclists \cite{liu2020deep}. LiDAR can provide high-fidelity 3D point cloud data to grasp the precise 3D location of the traffic objects~\cite{arnold2019survey}. RADAR sensor has been an integral part of safety-critical applications in automotive industry due to its robust performance on variable weather and lighting conditions~\cite{8443497}.

During the last couple of decades, a large portion of the object perception methods and high-fidelity perception data have came from on-board sensors while most of the roadside sensors are still used for traditional traffic data collection such as counting traffic volumes based on loop detectors, cameras, or radar~\cite{zou2019object}. Although empowered with advanced perception methods, on-board sensors are inevitably limited by range and occlusion of other objects. Infrastructure-based perception systems have the potential to achieve better object perception results with less occlusion effects and more flexibility in terms of mounting height and pose. 

In this survey paper, the infrastructure-based object detection and tracking methods are reviewed. This survey aims to establish an overall landscape for object perception based on roadside high-fidelity sensors and to provide inspirations for future research. The rest of this paper is organized as follows: Architectures for infrastructure-based perception system are reviewed in Section~\ref{sec:architecture} followed by the roadside sensors and general perception methodologies. Infrastructure-based object detection and tracking approaches are reviewed in Section~\ref{sec:Infra} followed by the Datasets and Simulators. The last section concludes this paper with further discussion.

\section{System Architectures}
\label{sec:architecture}
Defined by the Society of Automotive Engineers (SAE) J3216 Standard~\cite{2021SAE}, Cooperative Driving Automation (CDA) enables communication and cooperation between properly equipped vehicles, infrastructure, and other road users. Led by the Federal Highway Administration (FHWA), CARMA~\cite{2021CARMA} is one of the state-of-the-art (SOTA) programs for CDA.

\begin{figure}[!ht]
    \centering
    \includegraphics[width=0.5\textwidth]{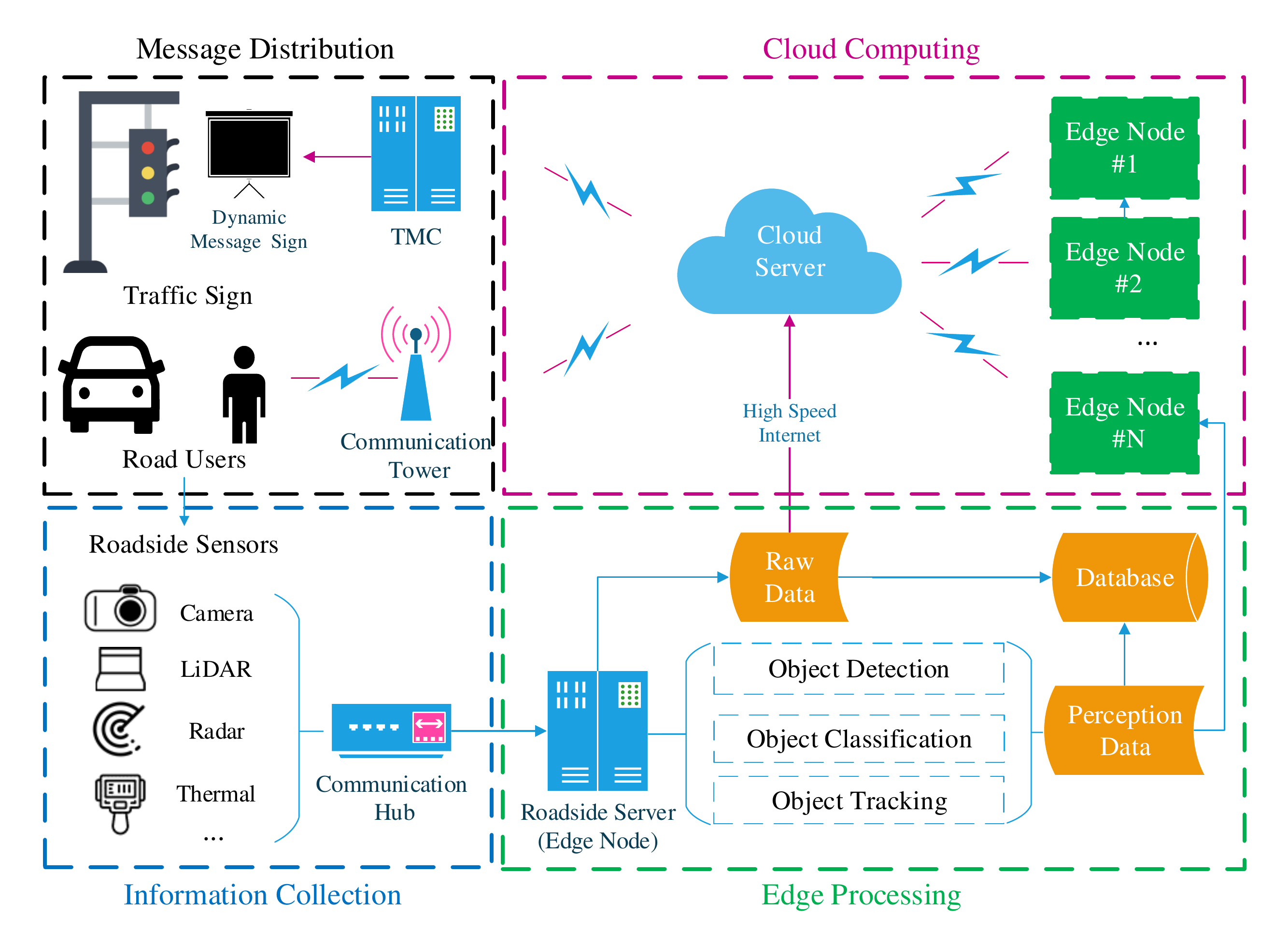}
    \caption{Visualization of systematic architecture for infrastructure-based perception system.}
    \label{fig:roadisde perception structure}
\end{figure}

According to the definition by SAE, CDA has four classes of cooperative automation with an increasing amount of cooperation associated with information shared among CDA participants. Hence, the fidelity and range of perception information have significant impact on the subsequent cooperation performance. Fig.~\ref{fig:roadisde perception structure} demonstrates a systematic architecture of infrastructure-based object perception system for enabling CDA. Specifically, four typical phases are identified in the infrastructure-based object perception process: 1) \textit{Information Collection}; 2) \textit{Edge Processing}; 3) \textit{Cloud Fusion}; and 4) \textit{Information Distribution}.
    
\subsection{Information Collection}
Traditional roadside sensors, such as \textit{Loop Detectors} and \textit{Microwave RADAR}, are widely used for providing the source data for traffic surveillance and dynamic traffic management~\cite{nellore2016survey}. However, the main capacity of these traditional sensors is to provide the presence of objects at certain locations. Thanks to the advancement in high-performance computation and machine learning,  high-resolution sensors (e.g., camera, LiDAR, etc) are able to provide object-level perception results, which are equipped on roadside infrastructures to perceive the environment, and transmit collected data to the roadside server via communication hub for further processing.

\subsection{Edge Processing}
Considering the limited bandwidth to transmit a large volume of raw data (e.g., point cloud datasets), information collected from roadside sensors may be processed on an edge (roadside) server. Generally, there are three main steps for processing the raw sensing data in this phase, as shown below:
\begin{itemize}
    \item \textit{Preprocessing}: Manipulations of raw data to provide a ready-to-use format for perception modules according to specific sensors, such as coordinate transformation, geo-fencing, and noise reduction.
    \item \textit{Object Perception}: Generation of object detection and tracking results for demonstrating position and pose, as well as identification of certain road users, such as rotated bounding boxes with unique ID and classification tag. Further, multi-sensor fusion algorithms may be applied if there are more than one sensors used for single perception node.
    \item \textit{Storage}: Recording of raw sensing data and perception data with timestamps for post-processing or analysis at edge-side.
\end{itemize}

\subsection{Cloud Computing}
Generally, perception data are generated by roadside equipment (RSE) due to the large volume of raw data and then the perception data will be transmitted to the \textit{Cloud} via wireless communication (e.g, \textit{Cellular Network}, \textit{WLAN}, etc). In some systems equipped with high-speed Internet to allow the high-volume low-latency data transmission, raw data could also be transmitted to the Cloud for processing. In terms of multi-node perception systems, i.e., simultaneously perceiving the environment from different locations, time alignment (with necessity of delay compensation) and object association need to be considered for spatiotemporal information assimilation and synchronization.

\subsection{Message Distribution}
The perception information (along with any advisory or actuation signals) can be distributed to road users in two major ways, depending on the connectivity status: for conventional road users without wireless connectivity, such information can be delivered to end devices at the roadside, such as \textit{Dynamic Message Sign} (DMS) or signal head display of traffic lights via the Traffic Management Center (TMC). For connected road users with wireless communications, customized information, e.g., surrounding objects and \textit{Signal Phase and Timing} (SPaT) of upcoming signals, can be accessed to enable various cooperative driving automation (CDA) applications, such as Cooperative Eco-Driving~\cite{altan2017glidepath, bai2022hybrid}.

\section{Roadside Sensors}
\label{sec: sensors}
For infrastructure-based object detection and tracking system, roadside sensors are the fundamental modules for data collection. This section overviews typical types of roadside sensors from different perspectives.
\subsection{Configuration and Performance}

\begin{table*}[!t]
\centering
\caption{Performance matrix for different sensors utilized for infrastructure-based perception.}
\label{tab:sensors}
\resizebox{\textwidth}{!}{%
\begin{tabular}{l|l|l|l|l|l|l}
\hline
\multicolumn{1}{c|}{Capabilities}             & \multicolumn{1}{c|}{Camera} & \multicolumn{1}{c|}{LiDAR} & \multicolumn{1}{c|}{RADAR} & \multicolumn{1}{c|}{Thermal} & \multicolumn{1}{c|}{Fisheye} & \multicolumn{1}{c}{Loop} \\ \hline
Privacy-safe data                             & $\star$                     & $\star\star\star$          & $\star\star\star$          & $\star\star$                 & $\star$                      & $\star\star\star$         \\
Accurately detects and classifies objects     & $\star\star$                & $\star\star\star$          & $\star$                    & $\star\star$                 & $\star\star$                 & $\star$              \\
Accurately measures object speed and position & $\star\star$                & $\star\star\star$          & $\star\star\star$          & $\star\star$                 & $\star\star$                      & $\star\star$         \\
Extensive field-of-view (FOV)  & $\star$                     & $\star\star\star$          & $\star\star$               & $\star$                      & $\star\star\star$            & $\star$                   \\
Reliability across changes in lighting, sun, temperature     & $\star\star$                & $\star\star\star$          & $\star\star\star$          & $\star\star$                 & $\star\star$                 & $\star\star\star$         \\
Ability to read signs and differentiate color & $\star\star\star$           & $\star\star$               & $\star$                    & $\star$                      & $\star\star\star$            & $\star$                   \\ \hline
\end{tabular}%
}
\end{table*}

Regarding the installation of roadside sensors, typical locations may include signal arm and street lamp post, with some minimum height requirement to avoid tampering. As a result, roadside sensors can have a much higher position (compared to on-board sensors) to minimize the occlusion effect due to dense traffic. The specific installation position may vary based on different roadside sensors. For example, the roadside LiDAR sensors are mainly installed at the height of 10 - 20 ft (but no more than 30 ft), while fisheye cameras prefer a higher installation.

For general performance of different sensors used in a roadside perception system, Table~\ref{tab:sensors} provides a summary on those that are widely utilized in roadside traffic surveillance. Each of these sensors has its own capabilities and strengths in different use cases. 

\subsection{Operational Pipeline}
\begin{figure}[!ht]
    \centering
    \includegraphics[width=0.5\textwidth]{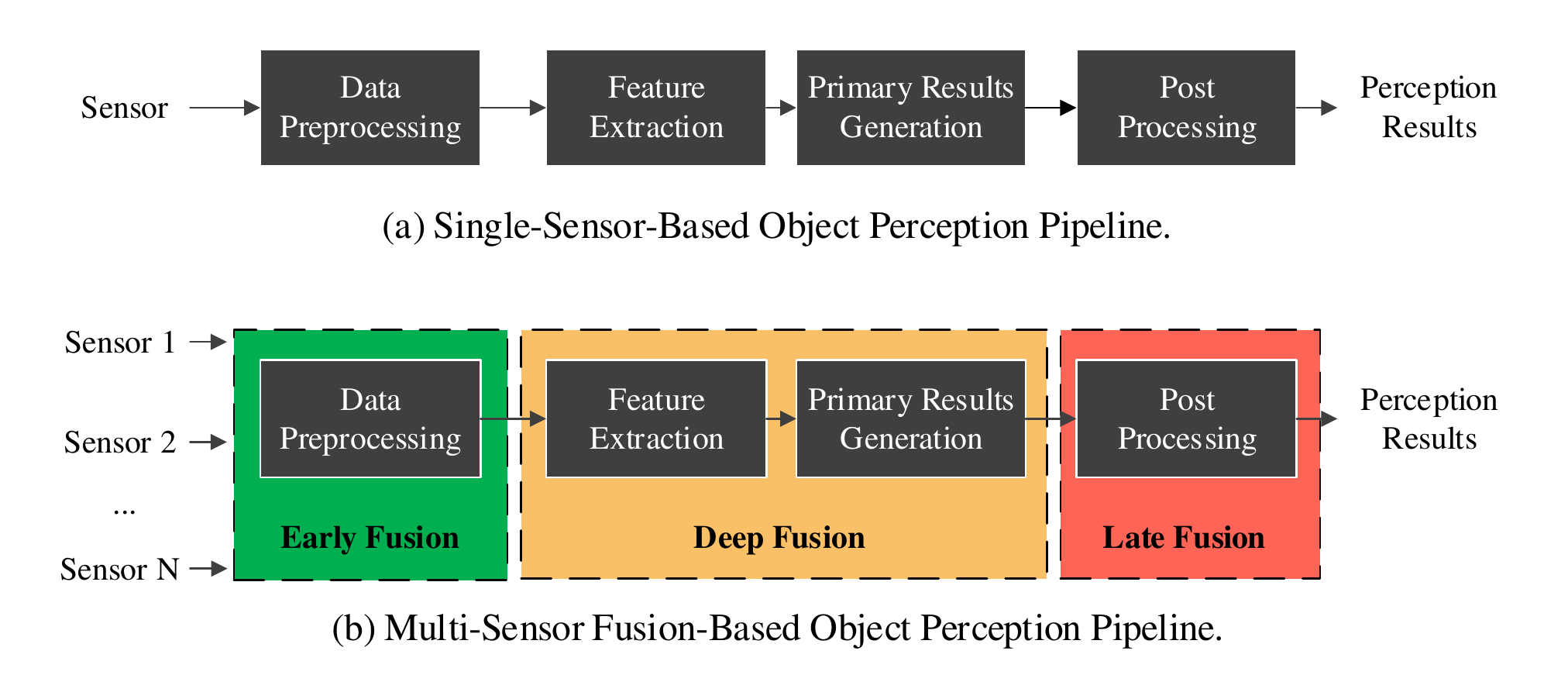}
    \caption{Systematic diagram of operational pipeline for: (a) single-sensor-based perception model; and (b) multi-sensor-based perception model.}
    \label{fig:pipelines}
\end{figure}

In terms of the number of sensors applied, the systematic operational pipeline of object detection and tracking based on roadside sensors can be divided into two main categories, i.e., single-sensor-based and multi-sensor-based, as shown in Fig.~\ref{fig:pipelines}.
\subsubsection{Single-sensor-based Perception}
Single-sensor-based object detection and tracking systems have been widely developed and applied in the real-world transportation system whose main pipeline is demonstrated in Fig.~\ref{fig:pipelines}~(a). Data collected from the sensor is first \textit{preprocessed} to reduce noise, filter unrelated data, and properly reformat for downstream modules. Then, \textit{feature extraction} is applied to calculate predefined features by mathematical models (if based on traditional methods) or to generate hidden features by neural networks (if based on deep learning). Detection and tracking results are generated by the \textit{perception} module and are fed into the \textit{post-processing} module to further clean the perception outputs (e.g., filtering overlapped bounding boxes and predictions with score under the threshold).
\subsubsection{Multi-sensor-based Perception}
Compared with single-sensor-based perception systems, multi-sensor-based perception systems have the potential to achieve better object detection and tracking performance via sensor fusion, owing to the complementary of different sensors. In terms of the stage of sensor fusion, multi-sensor perception system can be divided into three classes: 1) Early Fusion -- to fuse raw data at the preprocessing stage; 2) Deep Fusion -- to fuse features at the feature extraction stage; and 3) Late Fusion -- to fuse perception results at the post-processing stage. Different fusion schemes both have pros and cons in terms of different perspectives. For instance, Early Fusion and Deep Fusion have the strength for fusion accuracy but need more computational power and complex model design. Conversely, Late Fusion can achieve better real-time performance but will sacrifice accuracy. It depends on the specific demands under different traffic scenarios to determine the deployment of fusion schemes.

\section{General Perception Methodology}
\label{sec: General Perception Methodologies}
In this section, general perception methodologies which act as the building blocks for infrastructure-based perception methods, are briefly reviewed. However, due to the limited space, only several object detection milestones will be covered chronologically from two perspectives: the traditional approach and the deep-learning approach. 
\subsection{Traditional Approach}
Approximately 20 years ago, Viola and Jones~\cite{viola2001rapid, viola2004robust} proposed a method for real-time detection of human faces without any constraints. This algorithm outperformed any other contemporary algorithms in terms of real-time performance, without compromising detection accuracy. In 2005, Dalal and Triggs~\cite{dalal2005histograms} proposed the Histogram of Oriented Gradients (HOG) feature descriptor which provided significant improvement of the scale-invariant feature transform~\cite{lowe1999object, lowe2004distinctive} and shape context~\cite{belongie2002shape}. The HOG detector has been regarded as the cornerstone for many subsequent object detectors and implemented in various real-world applications~\cite{felzenszwalb2008discriminatively, felzenszwalb2010cascade, malisiewicz2011ensemble}. Deformable Part-based Model (DPM) proposed by Felzenszwalb~\cite{felzenszwalb2008discriminatively} consecutively won the Pascal Visual Object Classes (VOC)-07, -08, and -09 detection challenges~\cite{everingham2010pascal}. Due to their dominant performance, DPM and its variants~\cite{felzenszwalb2010cascade} are widely regarded as the pinnacle of traditional object detection methods~\cite{zou2019object}.

\subsection{Deep-Learning Approaches}
Benefiting from the increased computational power, convolutional neural networks (CNNs)~\cite{krizhevsky2012imagenet} started to be widely used in 2012. Two years later, Girshick et al. proposed the Regions with CNN features (R-CNN) for object detection and completely unfolded the advantage of deep learning~\cite{girshick2014rich, girshick2015region}. In the same year, Spatial Pyramid Pooling Networks (SPPNet) proposed by He et al. was able to generate feature representation regardless of the image size, and run 20 times faster than R-CNN without compromising accuracy~\cite{he2015spatial}. In 2015, multiple 
renowned detectors were proposed by researchers: 1) Fast R-CNN~\cite{girshick2015fast} -- over 200 times faster than R-CNN -- proposed by Girshick; 2) Faster R-CNN~\cite{ren2015faster, ren2016faster} -- the first end-to-end, and the first near-realtime deep learning detector -- proposed by Ren et al.; 3) You Only Look Once (YOLO)~\cite{redmon2016you} -- the first one-stage detector in the deep learning era with extremely fast speed (45 - 155 fps) -- proposed by Joseph et al.; and 4) Single Shot MultiBox Detector (SSD)~\cite{liu2016ssd} -- the second one-stage detector but with significantly improved accuracy -- proposed by Liu et al. In 2017, Lin et al. proposed Feature Pyramid Networks (FPN)~\cite{lin2017feature} based on Faster R-CNN, which achieved the SOTA object detection performance and has become a fundamental building block for various object perception models. In recent years, \textit{Transformers}~\cite{vaswani2017attention} embedded with the mechanism of attention has been leading the trend to the majority of object perception tasks, such as Vision Transformer (ViT) proposed by Dosovitskiy et al.~\cite{dosovitskiy2020vit}, and Swin-Transformer proposed by Liu et al~\cite{liu2021swin}.

\section{Infrastructure-based Object Detection and Tracking Approaches}
\label{sec:Infra}
Although general object detection has gone through a rapid development era, object detection and tracking based on roadside sensors is still an emerging topic and has the potential to break the current bottleneck for autonomous driving especially in a mixed traffic environment via cooperative perception~\cite{gupta2021deep}. This section reviews the infrastructure-based object detection and tracking approaches with analysis of details in literature. Since the camera-based perception work has been reviewed comprehensively in previous surveys~\cite{zou2019object,datondji2016survey}, this section will mainly focus on roadside LiDAR-based perception methods.

\subsection{2D Object Detection}

Roadside camera system have been widely used for object detection to support traffic surveillance, safety warning and various other applications. Ojala et al. proposed a CNN-based pedestrian detection and localization approach using roadside camera~\cite{Ojala8793228}. The perception system consists of a monovision camera streaming video and a computing unit which performs object detection and distance measurements on the detected objects.

Using a roadside LiDAR, Zhang et al.~\cite{zhang2020gc} proposed GC-net, a three-stage pipeline, including gridding, clustering, and classification. The raw point cloud data (PCD) is mapped into a grid structure and then clustered by the Grid-Density-Based Spatial Clustering algorithm. Finally, a CNN-based classifier is applied to categorize the detected objects by extracting the local features.

Liu et al. proposed a roadside LiDAR-based object detection approach by background filtering and clustering~\cite{Liu9434525}. Specifically, the background filtering method is designed on the basis of point correlation by KDTree~\cite{redmond2007method} neighborhood searching and the clustering is based on an adaptive (threshold) Euclidean clustering method.

Song et al. proposed a layer-based method for background filtering and object detection~\cite{Song9216093}. Specifically, a layer-based searching method is designed on the basis of feature distribution of PCD to distinguish moving objects from the point cloud. The Density-Based Spatial Clustering Applications with Noise (DBSCAN)~\cite{ester1996density} method can be applied for point clustering and generate the object detection results.

Gouda et al. proposed an automated approach to mapping and assessing roadside clearance parameters using LiDAR on rural highways~\cite{gouda2021automated}. Pavement edge trajectories are extracted based on the pavement surface point extracted from PCD. Then, a voxel-based raycasting approach is applied to search for roadside objects and query their locations, and non-compliant locations with substandard conditions are automatically queried.

Zhang et al. proposed an object detection method based on background construction and clustering from roadside LiDAR data~\cite{8484040, zhang2019automatic}. The discrete horizontal and vertical angular values are regarded as coordinates of pixels in digital images, and the farthest and mean distance of each azimuth are used to construct the background dataset. Next, a density-based spatial clustering method~\cite{tran2013revised} is applied to generate the object detection results.

For a multi-sensor system, Zhu et al. proposed \textit{Multi-Sensor Multi-Level Enhanced YOLO} (\textit{MME-YOLO}) for vehicle detection in traffic surveillance~\cite{zhu2021mme}. MME-YOLO consists of two tightly coupled structures: 
\begin{itemize}
    \item The enhanced inference head is empowered by attention-guided feature selection blocks and anchor-based/anchor-free ensemble head in terms of better generalization abilities in real-world scenarios.  
    \item The LiDAR-Image composite module is based on CBNet~\cite{liu2020cbnet} to cascade the multi-level feature maps from the LiDAR subnet to the image subnet, which strengthens the generalization of the detector in complex scenarios.
\end{itemize}
Owing to the above innovations, MME-YOLO can achieve better performance for vehicle detection compared with YOLOv3~\cite{farhadi2018yolov3} for roadside sensor data.

Bai et al.~\cite{bai2022cyber} proposed a deep-learning based real-time vehicle detection and reconstruction system from roadside LiDAR data. Specifically, CARLA simulator~\cite{dosovitskiy2017carla} is implemented for collecting training dataset, and ComplexYOLO model~\cite{simony2018complex} is applied and retrained for the object detection on the CARLA dataset. Finally, a co-simulation platform is designed and developed to provide vehicle detection and object-level reconstruction, which aims to empower subsequent CDA applications with readily retrieved authentic detection data.

Except for detecting general road users (e.g., vehicles and pedestrians), Chen et al. provided an innovative attempt for deer crossing-road detection by using roadside LiDAR~\cite{chen2019deer}. The main workflow is adopted from their previous works, which is represented by a "background filtering-clustering-classification" process~\cite{8484040, zhao2019detection, zhang2019vehicle}. Particularly, several different clustering algorithms, e.g., naive Bayes~\cite{murphy2006naive}, random forest~\cite{breiman2001random} and k-nearest neighbor (KNN)~\cite{li2002detection}, are applied for classifying deer, pedestrians and vehicles.

Cicek proposed a deep-learning based automated curbside parking spot detection approach through a roadside camera~\cite{cicek2021fully}. To identify the road boundaries, object detection and road segmentation methods are employed by utilizing \textit{FCN-VGG16} model~\cite{long2015fully} on the \textit{KITTI} dataset~\cite{Geiger2012CVPR} and \textit{Faster R-CNN}~\cite{ren2015faster} on \textit{MS-COCO} dataset~\cite{lin2014microsoft}, respectively. Then, a method is designed to differentiate parked vehicles from the moving ones and then give the guidance of nearest spot information to drivers.

\subsection{3D Object Detection}

Guo et al. proposed a 3D vehicle detection method based on monocular camera~\cite{9502706}, which consists of three steps: 1) clustering arbitrary object contours into linear equations; 2) estimating positions, orientations and dimensions of vehicles by applying K-means method; and 3) refining 3D detection results by maximizing a posterior probability. 

For pedestrian detection, Gong et al. proposed a roadside LiDAR-based real-time detection approach by combining traditional and deep learning algorithms~\cite{gong2021pedestrian}. Several techniques are designed to guarantee the real-time performance, including: application of Octree with region-of-interest (ROI) selection, and development of an improved Euclidean clustering algorithm with adaptive search radius. The roadside system is equipped with \textit{NVIDIA Jetson AGX Xavier}, achieving the inference time of 110 ms per frame.

Bai et al.~\cite{bai2021cmm} proposed a deep-learning based 3D object detection, tracking, and reconstruction system for the real-world implementation. The field operational system consists of three main parts: 1) 3D object detection by adopting PointPillar~\cite{lang2019pointpillars} for inference from roadside PCD data; 2) 3D multi-object tracking by improving DeepSORT~\cite{veeramani2018deepsort} to support 3D tracking; and 3) 3D reconstruction by geodetic transformation and real-time on-board GUI display.

For multi-sensor perception systems, Arnold et al. proposed a cooperative 3D object detection model by utilizing multiple depth cameras to mitigate the limitation from FOV of a single-sensor system~\cite{arnold2020cooperative}. For each camera, a depth image is projected to a pseudo-point-cloud data~\cite{hartley2000zisserman, glennie2010static}. Two sensor-fusion schemes are designed: early fusion and late fusion (see Fig.~\ref{fig:pipelines}), and adopted based on Voxelnet~\cite{zhou2018voxelnet}. The evaluation in a T-junction and a roundabout scenario in CARLA simulator~\cite{dosovitskiy2017carla} demonstrates that the proposed method can enlarge the detection coverage without compromising accuracy.

\subsection{Object Detection and Tracking}
Using roadside LiDAR, Zhao et al. proposed a detection and tracking approach for pedestrians and vehicles~\cite{zhao2019detection}. As one of the early studies utilizing roadside LiDAR for perception, a classical detection and tracking pipeline for PCD was designed. It mainly consists of:
\begin{itemize}
    \item \textit{Background Filtering}: To remove the laser points reflected from road surface or buildings by applying a statistics-based background filtering method~\cite{wu2017automatic}.
    \item \textit{Clustering}: To generate clusters for the laser points by implementing a DBSCAN method~\cite{ester1996density}.
    \item \textit{Classification}: To generate different labels for different traffic objects, such as vehicles and pedestrians, based on neural networks~\cite{li2012brief}.
    \item \textit{Tracking}: To identify the same object in continuous data frames by applying a discrete Kalman filter~\cite{welch1995introduction}.
\end{itemize}
Based on the aforementioned work, Cui et al. designed an automatic vehicle tracking system by considering vehicle detection and lane identification~\cite{cui2019automatic}. A real-world operational system is developed, which consists of a roadside LiDAR, an edge computer, a \textit{Dedicated Short-Range Communication} (\textit{DSRC}) \textit{Roadside Unit} (\textit{RSU}), a Wi-Fi router, a DSRC \textit{On-board Unit} (\textit{OBU}), and a \textit{Graphic User Interface} (\textit{GUI}).

Following the similar workflow, Zhang et al. proposed a vehicle tracking and speed estimation approach based on a roadside LiDAR~\cite{zhang2020vehicle}. Vehicle detection results are generated by the "\textit{Background Filtering-Clustering-Classification}" process. Then, a centroid-based tracking flow is implemented to obtain initial vehicle transformations, and the unscented Kalman Filter~\cite{julier2004unscented} and joint probabilistic data association filter~\cite{bar2009probabilistic} are adopted in the tracking flow. Finally, vehicle tracking is refined through an BEV-LiDAR-image matching process to improve the accuracy of estimated vehicle speeds.

To mitigate the occlusion impact in dense traffic, Zhang et al. proposed an adjacent-frame fusion method for vehicle detection and tracking approach with a roadside LiDAR~\cite{zhang2019vehicle}. Compared with previous research~\cite{zhao2019detection}, objects can be detected and tracked without object model extraction or bounding box description, which improves the perception performance with occlusion.

MultEYE~\cite{balamuralidhar2021multeye} is a monitoring system for real-time vehicle detection, tracking and speed estimation proposed by Balamuralidhar et al. Different from general roadside sensors equipped on signal poles or light poles, the data source of MultEYE comes from an Unmanned Aerial Vehicle (UAV) equipped with an embedded computer and a video camera. Inspired by the multi-task learning methodology, a segmentation head~\cite{paszke2016enet} is added to the object detector backbone~\cite{bochkovskiy2020yolov4}. Dedicated object tracking~\cite{bolme2010visual} and speed estimation algorithms have been optimized to track objects reliably from an UAV with limited computational efforts.


\section{Datasets and Simulators}
\subsection{General Datasets}
\subsubsection{General Object Detection Datasets}
Owing to prevailing needs in autonomous driving for surrounding perception, most datasets for object detection and tracking are collected from on-board sensors. Several widely used datasets for driving automation are briefly introduced as follows:
\begin{itemize}
    \item \textit{KITTI}: one of the most popular datasets, which consists of hours of traffic scenarios recorded with a variety of sensor modalities for mobile robotics and autonomous driving~\cite{Geiger2012CVPR}. 
    \item \textit{NuScenes}: the first dataset to carry the full autonomous vehicle sensor suite: 6 cameras, 5 radars and 1 LiDAR, all with full 360 degree field of view~\cite{caesar2020nuscenes}.
    \item \textit{Waymo Open Dataset}: a large-scale, high quality, diverse dataset which consists of 1150 scenes captured across a range of urban and suburban geography~\cite{sun2020scalability}.
\end{itemize}

\subsection{Roadside Datasets}
Because the roadside perception has great potential to promote the development of CDA, there are immediate demands of establishing a roadside sensor-based dataset for various infrastructure-based object perception tasks. In 2021, \textit{BAAI-VANJEE Roadside Dataset} was published by Deng et al. to support the Connected Automated Vehicle Highway technologies~\cite{yongqiang2021baai}. The BAAI-VANJEE Roadside Dataset consists of LiDAR data and RGB images collected by a roadside data-collection platform and contains 2500 frames of LiDAR data, 5000 frames of RGB images which includes 12 classes of objects, 74K 3D object annotations and 105K 2D object annotations.

\subsection{Simulators}
To promote the development of autonomous driving, some game engine-based simulators are developed to provide a cost-effective way for algorithm design and evaluation, such as CARLA~\cite{dosovitskiy2017carla}, SVL~\cite{rong2020lgsvl}, and AirSim~\cite{shah2018airsim}. These simulators are open-source with detailed tutorials and have the capabilities to provide high-resolution high-fidelity sensor data, such as various kinds of cameras and LiDARs. These simulators can provide a highly customized and cost-effective way for collecting training datasets and traffic scenarios, and thus are widely applied in learning-based object perception tasks~\cite{marvasti2020cooperative, arnold2020cooperative, arnold2019survey}.

\section{Discussion}
Although infrastructure-based object perception is an emerging research area, it is playing an increasingly significant role in promoting the perception capabilities for CDA applications. Many studies have been conducted to lay the foundation and provide inspirations for future work. In this section, we present our insights concerning the current state, open problems and future trends in infrastructure-based object detection and tracking for CDA applications.
\subsection{Current State and Open Challenges}
\subsubsection{Sensor System}
Limited by the space, this paper does not provide a comprehensive overview for every related approaches, but most of the studies mentioned are based on a single sensor, while few of them are proposed for multi-sensor perception system~\cite{zhu2021mme, arnold2020cooperative} in terms of roadside sensor-based perception. Some key open challenges are fusion schemes (e.g., early fusion, late fusion) of different sensor combinations and associated efficient fusion methods.
\subsubsection{Core Perception Methods}
According to the literature reviewed in Section~\ref{sec: General Perception Methodologies} and Section~\ref{sec:Infra}, there is an evident gap between general object perception and infrastructure-based object perception. For instance, the core methodologies of a large portion of the existing roadside LiDAR-based detection approaches are based on DBSCAN for clustering~\cite{Song9216093, 8484040, zhang2019automatic, zhao2019detection, zhang2020vehicle}, which has performance gap compared with the SOTA methods~\cite{lang2019pointpillars, zhou2018voxelnet}. Thus, one of the major challenges is the roadside data acquisition and annotation for promoting the deep learning-based research of infrastructure-based perception systems.

\subsubsection{Communications and Synchronization}
According to the experience from several field operational systems~\cite{cui2019automatic}, the synchronization problem caused by the processing and communication delay is a crucial issue for large-scale implementation.
\subsection{Future Trends}
\subsubsection{Towards Multi-Sensor Fusion}
A multi-sensor-based perception system has the potential to improve the perceiving performance by taking advantage of complementary sensor data~\cite{zamanakos2021comprehensive} with appropriate fusion techniques. An infrastructure-based perception system has more flexible conditions for multi-sensor equipment and is capable of empowering high-computational edge servers.
\subsubsection{Towards Cooperative Perception}
No matter how powerful the perception methods are, the physical occlusion could not be addressed by single-node perception. Perceiving the environment from multiple nodes has the capability to mitigate the limitation from occlusion, which is one of the major bottlenecks of current perception systems.

\subsubsection{Towards Lightweight OBU}
Although the on-board device has made major strides in development, it could be extremely costly to empower every single vehicle with a high-performance computation system for perception. Due to the rapid advance in high-speed wireless communication technologies, it becomes much more cost-effective to equip lightweight OBUs for local situation awareness perception and receive data from infrastructure-based high-performance nodes for wider range perception.

\section{Conclusions}
This paper provides an overview for infrastructure-based object detection and tracking systems. The architectures is presented to illustrate the four fundamental parts of infrastructure-based traffic object perception system. Roadside sensors are then introduced in terms of configuration, performance, operational pipelines, and general perception methodologies. Infrastructure-based object detection and tracking approaches are reviewed with detailed analyses followed by the brief introduction of datasets and simulators. Finally, this paper discusses current issues and future trends. To the best of our knowledge, this work is the first study that aims to provide a survey on the infrastructure-based traffic object detection and tracking methods.

\section*{Acknowledgments}
This research was funded by Toyota Motor North America, InfoTech Labs. The contents of this paper reflect the views of the authors, who are responsible for the facts and the accuracy of the data presented herein. The contents do not necessarily reflect the official views of the Toyota Motor North America.

\bibliographystyle{IEEEtran}  
\bibliography{references}  

\begin{thebibliography}{10}
\providecommand{\url}[1]{#1}
\csname url@samestyle\endcsname
\providecommand{\newblock}{\relax}
\providecommand{\bibinfo}[2]{#2}
\providecommand{\BIBentrySTDinterwordspacing}{\spaceskip=0pt\relax}
\providecommand{\BIBentryALTinterwordstretchfactor}{4}
\providecommand{\BIBentryALTinterwordspacing}{\spaceskip=\fontdimen2\font plus
\BIBentryALTinterwordstretchfactor\fontdimen3\font minus
  \fontdimen4\font\relax}
\providecommand{\BIBforeignlanguage}[2]{{%
\expandafter\ifx\csname l@#1\endcsname\relax
\typeout{** WARNING: IEEEtran.bst: No hyphenation pattern has been}%
\typeout{** loaded for the language `#1'. Using the pattern for}%
\typeout{** the default language instead.}%
\else
\language=\csname l@#1\endcsname
\fi
#2}}
\providecommand{\BIBdecl}{\relax}
\BIBdecl

\bibitem{2019Crash}
U.~D. of~Transportation, ``Overview of motor vehicle crashes in 2019,''
  Available:
  \url{https://crashstats.nhtsa.dot.gov/Api/Public/Publication/813060}, 2020.

\bibitem{2018Congestion}
INRIX, ``Inrix: Congestion costs each american 97 hours, \$1,348 a year,''
  Available: \url{https://inrix.com/press-releases/scorecard-2018-us/}, 2018.

\bibitem{2021Energy}
U.~D. of~Energy, ``Fotw \#1204: Fuel wasted due to u.s. traffic congestion in
  2020 cut in half from 2019 to 2020,'' Available:
  \url{https://www.energy.gov/eere/vehicles/articles/fotw-1204-sept-20-2021-fuel-wasted-due-us-traffic-congestion-2020-cut-half},
  2021.

\bibitem{fagnant2015preparing}
D.~J. Fagnant and K.~Kockelman, ``Preparing a nation for autonomous vehicles:
  opportunities, barriers and policy recommendations,'' \emph{Transportation
  Research Part A: Policy and Practice}, vol.~77, pp. 167--181, 2015.

\bibitem{2021SAE}
SAE, ``Taxonomy and definitions for terms related to cooperative driving
  automation for on-road motor vehicles j3216\_202005,'' Available:
  \url{https://www.sae.org/standards/content/j3216_202005/}, 2021.

\bibitem{wu2020improved}
J.~Wu, H.~Xu, Y.~Zhang, and R.~Sun, ``An improved vehicle-pedestrian near-crash
  identification method with a roadside lidar sensor,'' \emph{Journal of safety
  research}, vol.~73, pp. 211--224, 2020.

\bibitem{bai2022hybrid}
Z.~Bai, P.~Hao, W.~Shangguan, B.~Cai, and M.~J. Barth, ``Hybrid reinforcement
  learning-based eco-driving strategy for connected and automated vehicles at
  signalized intersections,'' \emph{IEEE Transactions on Intelligent
  Transportation Systems}, pp. 1--14, 2022.

\bibitem{wangCACC}
Z.~Wang, Y.~Bian, S.~E. Shladover, G.~Wu, S.~E. Li, and M.~J. Barth, ``A survey
  on cooperative longitudinal motion control of multiple connected and
  automated vehicles,'' \emph{IEEE Intelligent Transportation Systems
  Magazine}, vol.~12, no.~1, pp. 4--24, 2020.

\bibitem{liu2020deep}
L.~Liu, W.~Ouyang, X.~Wang, P.~Fieguth, J.~Chen, X.~Liu, and
  M.~Pietik{\"a}inen, ``Deep learning for generic object detection: A survey,''
  \emph{International journal of computer vision}, vol. 128, no.~2, pp.
  261--318, 2020.

\bibitem{arnold2019survey}
E.~Arnold, O.~Y. Al-Jarrah, M.~Dianati, S.~Fallah, D.~Oxtoby, and
  A.~Mouzakitis, ``A survey on 3d object detection methods for autonomous
  driving applications,'' \emph{IEEE Transactions on Intelligent Transportation
  Systems}, vol.~20, no.~10, pp. 3782--3795, 2019.

\bibitem{8443497}
A.~Manjunath, Y.~Liu, B.~Henriques, and A.~Engstle, ``Radar based object
  detection and tracking for autonomous driving,'' in \emph{2018 IEEE MTT-S
  International Conference on Microwaves for Intelligent Mobility (ICMIM)},
  2018, pp. 1--4.

\bibitem{zou2019object}
Z.~Zou, Z.~Shi, Y.~Guo, and J.~Ye, ``Object detection in 20 years: A survey,''
  \emph{arXiv preprint arXiv:1905.05055}, 2019.

\bibitem{2021CARMA}
F.~H.~A. USDoT, ``Carma,'' Available:
  \url{https://highways.dot.gov/tags/carma}, 2021.

\bibitem{nellore2016survey}
K.~Nellore and G.~P. Hancke, ``A survey on urban traffic management system
  using wireless sensor networks,'' \emph{Sensors}, vol.~16, no.~2, p. 157,
  2016.

\bibitem{altan2017glidepath}
O.~D. Altan, G.~Wu, M.~J. Barth, K.~Boriboonsomsin, and J.~A. Stark,
  ``Glidepath: Eco-friendly automated approach and departure at signalized
  intersections,'' \emph{IEEE Transactions on Intelligent Vehicles}, vol.~2,
  no.~4, pp. 266--277, 2017.

\bibitem{viola2001rapid}
P.~Viola and M.~Jones, ``Rapid object detection using a boosted cascade of
  simple features,'' in \emph{Proceedings of the 2001 IEEE computer society
  conference on computer vision and pattern recognition. CVPR 2001},
  vol.~1.\hskip 1em plus 0.5em minus 0.4em\relax Ieee, 2001, pp. I--I.

\bibitem{viola2004robust}
P.~Viola and M.~J. Jones, ``Robust real-time face detection,''
  \emph{International journal of computer vision}, vol.~57, no.~2, pp.
  137--154, 2004.

\bibitem{dalal2005histograms}
N.~Dalal and B.~Triggs, ``Histograms of oriented gradients for human
  detection,'' in \emph{2005 IEEE computer society conference on computer
  vision and pattern recognition (CVPR'05)}, vol.~1.\hskip 1em plus 0.5em minus
  0.4em\relax Ieee, 2005, pp. 886--893.

\bibitem{lowe1999object}
D.~G. Lowe, ``Object recognition from local scale-invariant features,'' in
  \emph{Proceedings of the seventh IEEE international conference on computer
  vision}, vol.~2.\hskip 1em plus 0.5em minus 0.4em\relax Ieee, 1999, pp.
  1150--1157.

\bibitem{lowe2004distinctive}
------, ``Distinctive image features from scale-invariant keypoints,''
  \emph{International journal of computer vision}, vol.~60, no.~2, pp. 91--110,
  2004.

\bibitem{belongie2002shape}
S.~Belongie, J.~Malik, and J.~Puzicha, ``Shape matching and object recognition
  using shape contexts,'' \emph{IEEE transactions on pattern analysis and
  machine intelligence}, vol.~24, no.~4, pp. 509--522, 2002.

\bibitem{felzenszwalb2008discriminatively}
P.~Felzenszwalb, D.~McAllester, and D.~Ramanan, ``A discriminatively trained,
  multiscale, deformable part model,'' in \emph{2008 IEEE conference on
  computer vision and pattern recognition}.\hskip 1em plus 0.5em minus
  0.4em\relax Ieee, 2008, pp. 1--8.

\bibitem{felzenszwalb2010cascade}
P.~F. Felzenszwalb, R.~B. Girshick, and D.~McAllester, ``Cascade object
  detection with deformable part models,'' in \emph{2010 IEEE Computer society
  conference on computer vision and pattern recognition}.\hskip 1em plus 0.5em
  minus 0.4em\relax Ieee, 2010, pp. 2241--2248.

\bibitem{malisiewicz2011ensemble}
T.~Malisiewicz, A.~Gupta, and A.~A. Efros, ``Ensemble of exemplar-svms for
  object detection and beyond,'' in \emph{2011 International conference on
  computer vision}.\hskip 1em plus 0.5em minus 0.4em\relax IEEE, 2011, pp.
  89--96.

\bibitem{everingham2010pascal}
M.~Everingham, L.~Van~Gool, C.~K. Williams, J.~Winn, and A.~Zisserman, ``The
  pascal visual object classes (voc) challenge,'' \emph{International journal
  of computer vision}, vol.~88, no.~2, pp. 303--338, 2010.

\bibitem{krizhevsky2012imagenet}
A.~Krizhevsky, I.~Sutskever, and G.~E. Hinton, ``Imagenet classification with
  deep convolutional neural networks,'' \emph{Advances in neural information
  processing systems}, vol.~25, pp. 1097--1105, 2012.

\bibitem{girshick2014rich}
R.~Girshick, J.~Donahue, T.~Darrell, and J.~Malik, ``Rich feature hierarchies
  for accurate object detection and semantic segmentation,'' in
  \emph{Proceedings of the IEEE conference on computer vision and pattern
  recognition}, 2014, pp. 580--587.

\bibitem{girshick2015region}
------, ``Region-based convolutional networks for accurate object detection and
  segmentation,'' \emph{IEEE transactions on pattern analysis and machine
  intelligence}, vol.~38, no.~1, pp. 142--158, 2015.

\bibitem{he2015spatial}
K.~He, X.~Zhang, S.~Ren, and J.~Sun, ``Spatial pyramid pooling in deep
  convolutional networks for visual recognition,'' \emph{IEEE transactions on
  pattern analysis and machine intelligence}, vol.~37, no.~9, pp. 1904--1916,
  2015.

\bibitem{girshick2015fast}
R.~Girshick, ``Fast r-cnn,'' in \emph{Proceedings of the IEEE international
  conference on computer vision}, 2015, pp. 1440--1448.

\bibitem{ren2015faster}
S.~Ren, K.~He, R.~Girshick, and J.~Sun, ``Faster r-cnn: Towards real-time
  object detection with region proposal networks,'' \emph{Advances in neural
  information processing systems}, vol.~28, pp. 91--99, 2015.

\bibitem{ren2016faster}
------, ``Faster r-cnn: towards real-time object detection with region proposal
  networks,'' \emph{IEEE transactions on pattern analysis and machine
  intelligence}, vol.~39, no.~6, pp. 1137--1149, 2016.

\bibitem{redmon2016you}
J.~Redmon, S.~Divvala, R.~Girshick, and A.~Farhadi, ``You only look once:
  Unified, real-time object detection,'' in \emph{Proceedings of the IEEE
  conference on computer vision and pattern recognition}, 2016, pp. 779--788.

\bibitem{liu2016ssd}
W.~Liu, D.~Anguelov, D.~Erhan, C.~Szegedy, S.~Reed, C.-Y. Fu, and A.~C. Berg,
  ``Ssd: Single shot multibox detector,'' in \emph{European conference on
  computer vision}.\hskip 1em plus 0.5em minus 0.4em\relax Springer, 2016, pp.
  21--37.

\bibitem{lin2017feature}
T.-Y. Lin, P.~Doll{\'a}r, R.~Girshick, K.~He, B.~Hariharan, and S.~Belongie,
  ``Feature pyramid networks for object detection,'' in \emph{Proceedings of
  the IEEE conference on computer vision and pattern recognition}, 2017, pp.
  2117--2125.

\bibitem{vaswani2017attention}
A.~Vaswani, N.~Shazeer, N.~Parmar, J.~Uszkoreit, L.~Jones, A.~N. Gomez,
  {\L}.~Kaiser, and I.~Polosukhin, ``Attention is all you need,'' in
  \emph{Advances in neural information processing systems}, 2017, pp.
  5998--6008.

\bibitem{dosovitskiy2020vit}
A.~Dosovitskiy, L.~Beyer, A.~Kolesnikov, D.~Weissenborn, X.~Zhai,
  T.~Unterthiner, M.~Dehghani, M.~Minderer, G.~Heigold, S.~Gelly, J.~Uszkoreit,
  and N.~Houlsby, ``An image is worth 16x16 words: Transformers for image
  recognition at scale,'' \emph{ICLR}, 2021.

\bibitem{liu2021swin}
Z.~Liu, Y.~Lin, Y.~Cao, H.~Hu, Y.~Wei, Z.~Zhang, S.~Lin, and B.~Guo, ``Swin
  transformer: Hierarchical vision transformer using shifted windows,''
  \emph{arXiv preprint arXiv:2103.14030}, 2021.

\bibitem{gupta2021deep}
A.~Gupta, A.~Anpalagan, L.~Guan, and A.~S. Khwaja, ``Deep learning for object
  detection and scene perception in self-driving cars: Survey, challenges, and
  open issues,'' \emph{Array}, p. 100057, 2021.

\bibitem{datondji2016survey}
S.~R.~E. Datondji, Y.~Dupuis, P.~Subirats, and P.~Vasseur, ``A survey of
  vision-based traffic monitoring of road intersections,'' \emph{IEEE
  transactions on intelligent transportation systems}, vol.~17, no.~10, pp.
  2681--2698, 2016.

\bibitem{Ojala8793228}
R.~Ojala, J.~Vepsäläinen, J.~Hanhirova, V.~Hirvisalo, and K.~Tammi, ``Novel
  convolutional neural network-based roadside unit for accurate pedestrian
  localisation,'' \emph{IEEE Transactions on Intelligent Transportation
  Systems}, vol.~21, no.~9, pp. 3756--3765, 2020.

\bibitem{zhang2020gc}
L.~Zhang, J.~Zheng, R.~Sun, and Y.~Tao, ``Gc-net: Gridding and clustering for
  traffic object detection with roadside lidar,'' \emph{IEEE Intelligent
  Systems}, 2020.

\bibitem{Liu9434525}
Z.~Liu, Q.~Li, S.~Mei, and M.~Huang, ``Background filtering and object
  detection with roadside lidar data,'' in \emph{2021 4th International
  Conference on Electron Device and Mechanical Engineering (ICEDME)}, 2021, pp.
  296--299.

\bibitem{redmond2007method}
S.~J. Redmond and C.~Heneghan, ``A method for initialising the k-means
  clustering algorithm using kd-trees,'' \emph{Pattern recognition letters},
  vol.~28, no.~8, pp. 965--973, 2007.

\bibitem{Song9216093}
Y.~Song, H.~Zhang, Y.~Liu, J.~Liu, H.~Zhang, and X.~Song, ``Background
  filtering and object detection with a stationary lidar using a layer-based
  method,'' \emph{IEEE Access}, vol.~8, pp. 184\,426--184\,436, 2020.

\bibitem{ester1996density}
M.~Ester, H.-P. Kriegel, J.~Sander, X.~Xu \emph{et~al.}, ``A density-based
  algorithm for discovering clusters in large spatial databases with noise.''
  in \emph{kdd}, vol.~96, no.~34, 1996, pp. 226--231.

\bibitem{gouda2021automated}
M.~Gouda, B.~Arantes~de Achilles~Mello, and K.~El-Basyouny, ``Automated object
  detection, mapping, and assessment of roadside clear zones using lidar
  data,'' \emph{Transportation research record}, vol. 2675, no.~12, pp.
  432--448, 2021.

\bibitem{8484040}
Z.~Zhang, J.~Zheng, X.~Wang, and X.~Fan, ``Background filtering and vehicle
  detection with roadside lidar based on point association,'' in \emph{2018
  37th Chinese Control Conference (CCC)}, 2018, pp. 7938--7943.

\bibitem{zhang2019automatic}
Z.~Zhang, J.~Zheng, H.~Xu, X.~Wang, X.~Fan, and R.~Chen, ``Automatic background
  construction and object detection based on roadside lidar,'' \emph{IEEE
  Transactions on Intelligent Transportation Systems}, vol.~21, no.~10, pp.
  4086--4097, 2019.

\bibitem{tran2013revised}
T.~N. Tran, K.~Drab, and M.~Daszykowski, ``Revised dbscan algorithm to cluster
  data with dense adjacent clusters,'' \emph{Chemometrics and Intelligent
  Laboratory Systems}, vol. 120, pp. 92--96, 2013.

\bibitem{zhu2021mme}
J.~Zhu, X.~Li, P.~Jin, Q.~Xu, Z.~Sun, and X.~Song, ``Mme-yolo: Multi-sensor
  multi-level enhanced yolo for robust vehicle detection in traffic
  surveillance,'' \emph{Sensors}, vol.~21, no.~1, p.~27, 2021.

\bibitem{liu2020cbnet}
Y.~Liu, Y.~Wang, S.~Wang, T.~Liang, Q.~Zhao, Z.~Tang, and H.~Ling, ``Cbnet: A
  novel composite backbone network architecture for object detection,'' in
  \emph{Proceedings of the AAAI conference on artificial intelligence},
  vol.~34, no.~07, 2020, pp. 11\,653--11\,660.

\bibitem{farhadi2018yolov3}
A.~Farhadi and J.~Redmon, ``Yolov3: An incremental improvement,'' in
  \emph{Computer Vision and Pattern Recognition}.\hskip 1em plus 0.5em minus
  0.4em\relax Springer Berlin/Heidelberg, Germany, 2018, pp. 1804--2767.

\bibitem{bai2022cyber}
Z.~Bai, G.~Wu, X.~Qi, K.~Oguchi, and M.~J. Barth, ``Cyber mobility mirror for
  enabling cooperative driving automation: A co-simulation platform,''
  \emph{arXiv preprint arXiv:2201.09463}, 2022.

\bibitem{dosovitskiy2017carla}
A.~Dosovitskiy, G.~Ros, F.~Codevilla, A.~Lopez, and V.~Koltun, ``Carla: An open
  urban driving simulator,'' in \emph{Conference on robot learning}.\hskip 1em
  plus 0.5em minus 0.4em\relax PMLR, 2017, pp. 1--16.

\bibitem{simony2018complex}
M.~Simony, S.~Milzy, K.~Amendey, and H.-M. Gross, ``Complex-yolo: An
  euler-region-proposal for real-time 3d object detection on point clouds,'' in
  \emph{Proceedings of the European Conference on Computer Vision (ECCV)
  Workshops}, 2018, pp. 0--0.

\bibitem{chen2019deer}
J.~Chen, H.~Xu, J.~Wu, R.~Yue, C.~Yuan, and L.~Wang, ``Deer crossing road
  detection with roadside lidar sensor,'' \emph{Ieee Access}, vol.~7, pp.
  65\,944--65\,954, 2019.

\bibitem{zhao2019detection}
J.~Zhao, H.~Xu, H.~Liu, J.~Wu, Y.~Zheng, and D.~Wu, ``Detection and tracking of
  pedestrians and vehicles using roadside lidar sensors,'' \emph{Transportation
  research part C: emerging technologies}, vol. 100, pp. 68--87, 2019.

\bibitem{zhang2019vehicle}
Z.~Zhang, J.~Zheng, H.~Xu, and X.~Wang, ``Vehicle detection and tracking in
  complex traffic circumstances with roadside lidar,'' \emph{Transportation
  research record}, vol. 2673, no.~9, pp. 62--71, 2019.

\bibitem{murphy2006naive}
K.~P. Murphy \emph{et~al.}, ``Naive bayes classifiers,'' \emph{University of
  British Columbia}, vol.~18, no.~60, pp. 1--8, 2006.

\bibitem{breiman2001random}
L.~Breiman, ``Random forests,'' \emph{Machine learning}, vol.~45, no.~1, pp.
  5--32, 2001.

\bibitem{li2002detection}
D.~Li, K.~D. Wong, Y.~H. Hu, and A.~M. Sayeed, ``Detection, classification, and
  tracking of targets,'' \emph{IEEE signal processing magazine}, vol.~19,
  no.~2, pp. 17--29, 2002.

\bibitem{cicek2021fully}
E.~Cicek and S.~G{\"o}ren, ``Fully automated roadside parking spot detection in
  real time with deep learning,'' \emph{Concurrency and Computation: Practice
  and Experience}, vol.~33, no.~23, p. e6006, 2021.

\bibitem{long2015fully}
J.~Long, E.~Shelhamer, and T.~Darrell, ``Fully convolutional networks for
  semantic segmentation,'' in \emph{Proceedings of the IEEE conference on
  computer vision and pattern recognition}, 2015, pp. 3431--3440.

\bibitem{Geiger2012CVPR}
A.~Geiger, P.~Lenz, and R.~Urtasun, ``Are we ready for autonomous driving? the
  kitti vision benchmark suite,'' in \emph{Conference on Computer Vision and
  Pattern Recognition (CVPR)}, 2012.

\bibitem{lin2014microsoft}
T.-Y. Lin, M.~Maire, S.~Belongie, J.~Hays, P.~Perona, D.~Ramanan,
  P.~Doll{\'a}r, and C.~L. Zitnick, ``Microsoft coco: Common objects in
  context,'' in \emph{European conference on computer vision}.\hskip 1em plus
  0.5em minus 0.4em\relax Springer, 2014, pp. 740--755.

\bibitem{9502706}
E.~Guo, Z.~Chen, S.~Rahardja, and J.~Yang, ``3d detection and pose estimation
  of vehicle in cooperative vehicle infrastructure system,'' \emph{IEEE Sensors
  Journal}, vol.~21, no.~19, pp. 21\,759--21\,771, 2021.

\bibitem{gong2021pedestrian}
Z.~Gong, Z.~Wang, B.~Zhou, W.~Liu, and P.~Liu, ``Pedestrian detection method
  based on roadside light detection and ranging,'' \emph{SAE International
  Journal of Connected and Automated Vehicles}, vol.~4, no. 12-04-04-0031,
  2021.

\bibitem{bai2021cmm}
Z.~Bai, S.~P. Nayak, X.~Zhao, G.~Wu, M.~J. Barth, X.~Qi, Y.~Liu, and K.~Oguchi,
  ``Cyber mobility mirror: Deep learning-based real-time 3d object perception
  and reconstruction using roadside lidar,'' \emph{arXiv preprint
  arXiv:2202.13505}, 2022.

\bibitem{lang2019pointpillars}
A.~H. Lang, S.~Vora, H.~Caesar, L.~Zhou, J.~Yang, and O.~Beijbom,
  ``Pointpillars: Fast encoders for object detection from point clouds,'' in
  \emph{Proceedings of the IEEE/CVF Conference on Computer Vision and Pattern
  Recognition}, 2019, pp. 12\,697--12\,705.

\bibitem{veeramani2018deepsort}
B.~Veeramani, J.~W. Raymond, and P.~Chanda, ``Deepsort: deep convolutional
  networks for sorting haploid maize seeds,'' \emph{BMC bioinformatics},
  vol.~19, no.~9, pp. 1--9, 2018.

\bibitem{arnold2020cooperative}
E.~Arnold, M.~Dianati, R.~de~Temple, and S.~Fallah, ``Cooperative perception
  for 3d object detection in driving scenarios using infrastructure sensors,''
  \emph{IEEE Transactions on Intelligent Transportation Systems}, 2020.

\bibitem{hartley2000zisserman}
R.~Hartley, ``A. zisserman multiple view geometry in computer vision,'' 2000.

\bibitem{glennie2010static}
C.~Glennie and D.~D. Lichti, ``Static calibration and analysis of the velodyne
  hdl-64e s2 for high accuracy mobile scanning,'' \emph{Remote sensing},
  vol.~2, no.~6, pp. 1610--1624, 2010.

\bibitem{zhou2018voxelnet}
Y.~Zhou and O.~Tuzel, ``Voxelnet: End-to-end learning for point cloud based 3d
  object detection,'' in \emph{Proceedings of the IEEE conference on computer
  vision and pattern recognition}, 2018, pp. 4490--4499.

\bibitem{wu2017automatic}
J.~Wu, H.~Xu, and J.~Zheng, ``Automatic background filtering and lane
  identification with roadside lidar data,'' in \emph{2017 IEEE 20th
  International Conference on Intelligent Transportation Systems (ITSC)}.\hskip
  1em plus 0.5em minus 0.4em\relax IEEE, 2017, pp. 1--6.

\bibitem{li2012brief}
J.~Li, J.-h. Cheng, J.-y. Shi, and F.~Huang, ``Brief introduction of back
  propagation (bp) neural network algorithm and its improvement,'' in
  \emph{Advances in computer science and information engineering}.\hskip 1em
  plus 0.5em minus 0.4em\relax Springer, 2012, pp. 553--558.

\bibitem{welch1995introduction}
G.~Welch, G.~Bishop \emph{et~al.}, ``An introduction to the kalman filter,''
  1995.

\bibitem{cui2019automatic}
Y.~Cui, H.~Xu, J.~Wu, Y.~Sun, and J.~Zhao, ``Automatic vehicle tracking with
  roadside lidar data for the connected-vehicles system,'' \emph{IEEE
  Intelligent Systems}, vol.~34, no.~3, pp. 44--51, 2019.

\bibitem{zhang2020vehicle}
J.~Zhang, W.~Xiao, B.~Coifman, and J.~P. Mills, ``Vehicle tracking and speed
  estimation from roadside lidar,'' \emph{IEEE Journal of Selected Topics in
  Applied Earth Observations and Remote Sensing}, vol.~13, pp. 5597--5608,
  2020.

\bibitem{julier2004unscented}
S.~J. Julier and J.~K. Uhlmann, ``Unscented filtering and nonlinear
  estimation,'' \emph{Proceedings of the IEEE}, vol.~92, no.~3, pp. 401--422,
  2004.

\bibitem{bar2009probabilistic}
Y.~Bar-Shalom, F.~Daum, and J.~Huang, ``The probabilistic data association
  filter,'' \emph{IEEE Control Systems Magazine}, vol.~29, no.~6, pp. 82--100,
  2009.

\bibitem{balamuralidhar2021multeye}
N.~Balamuralidhar, S.~Tilon, and F.~Nex, ``Multeye: Monitoring system for
  real-time vehicle detection, tracking and speed estimation from uav imagery
  on edge-computing platforms,'' \emph{Remote Sensing}, vol.~13, no.~4, p. 573,
  2021.

\bibitem{paszke2016enet}
A.~Paszke, A.~Chaurasia, S.~Kim, and E.~Culurciello, ``Enet: A deep neural
  network architecture for real-time semantic segmentation,'' \emph{arXiv
  preprint arXiv:1606.02147}, 2016.

\bibitem{bochkovskiy2020yolov4}
A.~Bochkovskiy, C.-Y. Wang, and H.-Y.~M. Liao, ``Yolov4: Optimal speed and
  accuracy of object detection,'' \emph{arXiv preprint arXiv:2004.10934}, 2020.

\bibitem{bolme2010visual}
D.~S. Bolme, J.~R. Beveridge, B.~A. Draper, and Y.~M. Lui, ``Visual object
  tracking using adaptive correlation filters,'' in \emph{2010 IEEE computer
  society conference on computer vision and pattern recognition}.\hskip 1em
  plus 0.5em minus 0.4em\relax IEEE, 2010, pp. 2544--2550.

\bibitem{caesar2020nuscenes}
H.~Caesar, V.~Bankiti, A.~H. Lang, S.~Vora, V.~E. Liong, Q.~Xu, A.~Krishnan,
  Y.~Pan, G.~Baldan, and O.~Beijbom, ``nuscenes: A multimodal dataset for
  autonomous driving,'' in \emph{Proceedings of the IEEE/CVF conference on
  computer vision and pattern recognition}, 2020, pp. 11\,621--11\,631.

\bibitem{sun2020scalability}
P.~Sun, H.~Kretzschmar, X.~Dotiwalla, A.~Chouard, V.~Patnaik, P.~Tsui, J.~Guo,
  Y.~Zhou, Y.~Chai, B.~Caine \emph{et~al.}, ``Scalability in perception for
  autonomous driving: Waymo open dataset,'' in \emph{Proceedings of the
  IEEE/CVF Conference on Computer Vision and Pattern Recognition}, 2020, pp.
  2446--2454.

\bibitem{yongqiang2021baai}
D.~Yongqiang, W.~Dengjiang, C.~Gang, M.~Bing, G.~Xijia, W.~Yajun, L.~Jianchao,
  F.~Yanming, and L.~Juanjuan, ``Baai-vanjee roadside dataset: Towards the
  connected automated vehicle highway technologies in challenging environments
  of china,'' \emph{arXiv preprint arXiv:2105.14370}, 2021.

\bibitem{rong2020lgsvl}
G.~Rong, B.~H. Shin, H.~Tabatabaee, Q.~Lu, S.~Lemke, M.~Mo{\v{z}}eiko,
  E.~Boise, G.~Uhm, M.~Gerow, S.~Mehta \emph{et~al.}, ``Lgsvl simulator: A high
  fidelity simulator for autonomous driving,'' \emph{arXiv preprint
  arXiv:2005.03778}, 2020.

\bibitem{shah2018airsim}
S.~Shah, D.~Dey, C.~Lovett, and A.~Kapoor, ``Airsim: High-fidelity visual and
  physical simulation for autonomous vehicles,'' in \emph{Field and service
  robotics}.\hskip 1em plus 0.5em minus 0.4em\relax Springer, 2018, pp.
  621--635.

\bibitem{marvasti2020cooperative}
E.~E. Marvasti, A.~Raftari, A.~E. Marvasti, Y.~P. Fallah, R.~Guo, and H.~Lu,
  ``Cooperative lidar object detection via feature sharing in deep networks,''
  in \emph{2020 IEEE 92nd Vehicular Technology Conference
  (VTC2020-Fall)}.\hskip 1em plus 0.5em minus 0.4em\relax IEEE, 2020, pp. 1--7.

\bibitem{zamanakos2021comprehensive}
G.~Zamanakos, L.~Tsochatzidis, A.~Amanatiadis, and I.~Pratikakis, ``A
  comprehensive survey of lidar-based 3d object detection methods with deep
  learning for autonomous driving,'' \emph{Computers \& Graphics}, vol.~99, pp.
  153--181, 2021.

\end{thebibliography}

\end{document}